\documentclass[letterpaper, 10 pt, conference]{ieeeconf}  %

\IEEEoverridecommandlockouts                              %

\usepackage{cite}
\usepackage{amsmath,amssymb,amsfonts}
\usepackage{algorithmic}
\usepackage{graphicx}
\usepackage{textcomp}
\usepackage{xcolor}

\usepackage{comment}

\usepackage{xcolor}
\usepackage{pagecolor}
\usepackage{lipsum}  
\usepackage{mdframed}

\definecolor{tumblue}{RGB}{0,101,189}
\definecolor{pantone301}{RGB}{0,82,147}
\definecolor{pantone540}{RGB}{0,51,89}
\definecolor{orangetum}{RGB}{227,114,34}

\def\BibTeX{{\rm B\kern-.05em{\sc i\kern-.025em b}\kern-.08em
    T\kern-.1667em\lower.7ex\hbox{E}\kern-.125emX}}

\definecolor{TUMBlue}{RGB}{0,101,189}%
\definecolor{TUMWhite}{RGB}{255,255,255}%
\definecolor{TUMBlack}{RGB}{0,0,0}%
\definecolor{TUMBlue1}{RGB}{0,51,89}%
\definecolor{TUMBlue2}{RGB}{0,82,147}%
\definecolor{TUMGray1}{RGB}{51,51,51}%
\definecolor{TUMGray2}{RGB}{127,127,127}%
\definecolor{TUMGray3}{RGB}{204,204,204}%
\definecolor{TUMBlue3}{RGB}{100,160,200}%
\definecolor{TUMBlue4}{RGB}{152,198,234}%
\definecolor{TUMIvory}{RGB}{218,215,203}%
\definecolor{TUMOrange}{RGB}{227,114,34}%
\definecolor{TUMGreen}{RGB}{162,173,0}%

\usepackage[T1]{fontenc}

\usepackage{tikz}
\usepackage{pgfplots}
\usepackage[per-mode=fraction, detect-weight=true]{siunitx}
\usepackage[hidelinks]{hyperref}
\usepackage[inkscapelatex=false, inkscapepath=./build/svg-inkscape]{svg}

\usepgfplotslibrary{groupplots}

\definecolor{darkgray}{RGB}{32, 32, 32}

\pgfplotsset{
	compat=1.18,
	standard-plot/.style={
			enlarge x limits=0,
			const plot,
			grid=major,
			grid style={dotted},
		},
	line-plot/.style={
			standard-plot,
			width=\columnwidth,
			height=0.5\columnwidth,
			every tick label/.append style={font=\footnotesize},
			label style={font=\footnotesize},
		},
	xy-plot/.style={
			width=\columnwidth,
			height=0.5\columnwidth,
			standard-plot,
		},
	group-plot/.style={
			width=0.975\columnwidth,
			height=0.38\columnwidth,
			every tick label/.append style={font=\footnotesize}
		},
	group-plot-2/.style={
			group-plot,
			group style={
					group size=1 by 2,
					xlabels at=edge bottom,
					xticklabels at=edge bottom,
					vertical sep=0cm,
				},
		},
	group-plot-3/.style={
			group-plot,
			group style={
					group size=1 by 3,
					xlabels at=edge bottom,
					xticklabels at=edge bottom,
					vertical sep=0cm,
				},
		},
	group-plot-4/.style={
			group-plot,
			group style={group size=1 by 3,
					xlabels at=edge bottom,
					xticklabels at=edge bottom,
					vertical sep=0.0cm,
				},
		},
	first-group-plot/.style={
			standard-plot,
			enlarge x limits=0,
			enlarge y limits=0.2,
			xmin=0,
			legend style={
					at={(0.99, 0.5)},anchor=east,
					nodes={scale=0.8, transform shape}
				},
			y label style={
				},
			label style={font=\footnotesize},
		},
	middle-group-plot/.style={
			standard-plot,
			enlarge x limits=0,
			enlarge y limits=0.2,
			xmin=0,
			legend style={
					at={(0.99, 0.5)},anchor=east,
					nodes={scale=0.8, transform shape}
				},
			y label style={
				},
			label style={font=\footnotesize},
		},
	last-group-plot/.style={
			standard-plot,
			xlabel=$t$ in \si{\second},
			enlarge x limits=0,
			enlarge y limits=0.2,
			xtick={0,0.04, 0.08, 0.12, 0.16, 0.2},
			xticklabels={0,0.04, 0.08, 0.12, 0.16, 0.2},
			xmin=0,
			legend style={
					at={(0.99, 0.5)},anchor=east,
					nodes={scale=0.8, transform shape}
				},
			y label style={
				},
			label style={font=\footnotesize},
		},
	generic-linestyle/.style={thick, mark=*, mark size=1.0pt},
	reference/.style={generic-linestyle, TUMBlack},
	line-1/.style={generic-linestyle, TUMBlue2},
	line-2/.style={generic-linestyle, black},
	line-3/.style={generic-linestyle, TUMGreen},
}

\usetikzlibrary{shapes}
\usetikzlibrary{positioning}
\usetikzlibrary{pgfplots.statistics}

\tikzset{
	flow-chart/.style={
			fc-block/.style={minimum height=1cm, minimum width=3.5cm, draw=TUMBlack, align=center},
			process/.style={fc-block, rectangle},
			process-predef/.style={fc-block, predproc},
			decision/.style={fc-block, diamond, aspect=2},
			flow-fusion/.style={decision, minimum width=1cm, aspect =1},
			input-output/.style={fc-block, trapezium, trapezium left angle = 65,trapezium right angle = 115,trapezium stretches},
			termination/.style={fc-block, rounded rectangle},
			flow/.style={-latex},
			flow-label/.style={fill=white, pos=0.5}
		},
	simulation-architecture/.style = {
			standard-block/.style={
					rectangle,
					rounded corners=.2cm,
					minimum height=1cm,
					minimum width=2cm,
					draw=TUMBlack,
					align=center,
				},
			simulation-block/.style={
					standard-block,
				},
			software-block/.style={
					standard-block,
					draw=TUMBlue
				},
			software-block/.style={
					standard-block,
					white,
					draw=TUMBlue,
					fill=TUMBlue,
				},
			data-block/.style={
					standard-block,
				},
			standard-arrow/.style={-latex},
			standard-arrow-label/.style={fill=none, pos=0.5, align=left}
		}
}

\newcommand{\etal}[1]{#1 et al.}

\newcommand{\githuburl}{https://github.com/TUMFTM/TAM__common}

\newcommand{\ros}{ROS~2}
\newcommand{\statestale}{\texttt{STALE}}
\newcommand{\stateerror}{\texttt{ERROR}}
\newcommand{\statewarn}{\texttt{WARN}}
\newcommand{\stateok}{\texttt{OK}}

\newcommand{\esser}[1]{#1}
\newcommand{\ebner}[1]{#1}
\newcommand{\betz}[1]{#1}

\DeclareSIUnit\bar{bar}

\usepackage[T1]{fontenc}

\usepackage{tikz}
\usepackage{pgfplots}
\usepackage[per-mode=fraction, detect-weight=true]{siunitx}
\usepackage[hidelinks]{hyperref}
\usepackage[inkscapelatex=false, inkscapepath=./build/svg-inkscape]{svg}

\usepgfplotslibrary{groupplots}

\title{\LARGE \bf Approaching Current Challenges in Developing a Software Stack for Fully Autonomous Driving}

\author{Simon Sagmeister, Simon Hoffmann, Tobias Betz, Dominic Ebner, Daniel Esser and Markus Lienkamp%
\thanks{*This work was funded by the Deutsche Forschungsgemeinschaft (DFG, German Research Foundation) – 469341384}%
\thanks{All authors are with the Technical University of Munich, Germany; School of Engineering \& Design, Department of Mobility Systems Engineering, Institute of Automotive Technology\newline{}
Corresponding author: \href{mailto:simon.sagmeister@tum.de}{simon.sagmeister@tum.de}}
}

\newcommand\copyrighttext{%
	\footnotesize \textcopyright 2025 IEEE.  Personal use of this material is permitted.  Permission from IEEE must be obtained for all other uses, in any current or future media, including reprinting/republishing this material for advertising or promotional purposes, creating new collective works, for resale or redistribution to servers or lists, or reuse of any copyrighted component of this work in other works.
}

\newcommand\copyrightnotice{%
	\begin{tikzpicture}[remember picture,overlay]
	\node[anchor=south,yshift=10pt, xshift=0pt] at (current page.south) {\fbox{\parbox{\dimexpr\textwidth-\fboxsep-\fboxrule\relax}{\copyrighttext}}};
	\end{tikzpicture}%
    \vspace{-0.35cm}
}

\begin{document}

\bstctlcite{IEEEexample:BSTcontrol}

\maketitle
\thispagestyle{empty}
\pagestyle{empty}
\copyrightnotice{}

\setcounter{footnote}{0}
\begin{abstract}
    Autonomous driving is a complex undertaking. A common approach is to break down the driving task into individual subtasks through modularization. These sub-modules are usually developed and published separately. However, if these individually developed algorithms have to be combined again to form a full-stack autonomous driving software, this poses particular challenges.
   Drawing upon our practical experience in developing the software of TUM Autonomous Motorsport, we have identified and derived these challenges in developing an autonomous driving software stack within a scientific environment. We do not focus on the specific challenges of individual algorithms but on the general difficulties that arise when deploying research algorithms on real-world test vehicles.
    To overcome these challenges, we introduce strategies that have been effective in our development approach. We additionally provide open-source implementations that enable these concepts on GitHub.
    As a result, this paper's contributions will simplify future full-stack autonomous driving projects, which are essential for a thorough evaluation of the individual algorithms.
\end{abstract}

\section{Introduction}

Autonomous driving is a rapidly growing research field with a continuously increasing number of publications~\cite{hacohen2022}.
However, only a few algorithms and approaches have been deployed in real-world applications.

A recent review paper~\cite{betz2022} shows that out of 111 publications in the area of motion planning and vehicle control,
only \SI{44}{\percent} have been deployed in an autonomous driving software stack. However, standalone testing of individual modules is insufficient for a modular approach, as the performance of each module is interdependent with other modules in the software stack~\cite{baumann2024}. Considering real-world experiments, \etal{Betz}~\cite{betz2022} show that only \SI{15}{\percent} out of these 111
racing-focused publications have been evaluated on a full-scale vehicle.
Comparing different algorithms becomes more challenging in the absence of real-world validation, as assessing an algorithm's performance is difficult without an actual test vehicle, particularly for certain types of algorithms~\cite{baumann2024}.
We obtained these numbers from publications on autonomous racing. We do not know the exact numbers for publications on autonomous driving in general. However, autonomous racing often takes place in competitions involving a real test vehicle, therefore potentially inflating the percentage of deployed algorithms compared to autonomous driving on public roads.

\begin{figure}[!tb]
    \centering
    \vspace*{0.2cm}
    \includegraphics[width=\columnwidth]{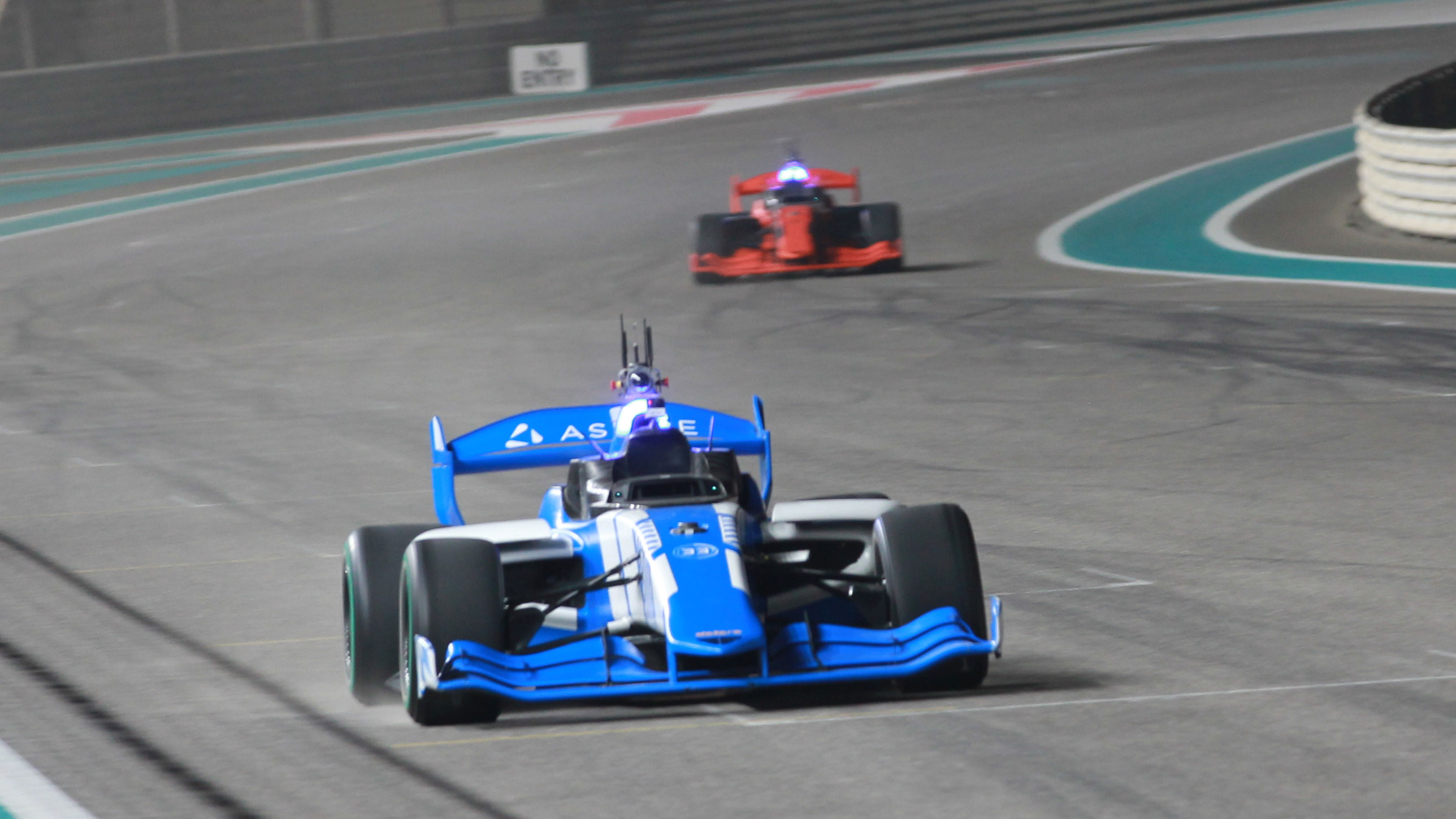}
    \caption{The EAV24 racing vehicle of TUM Autonomous Motorsport used during the Abu Dhabi Autonomous Racing League. It is based on Dallara's SF23 Super Formula Chassis and contains an autonomous driving hardware stack in the space usually occupied by the driver.}
    \label{fig:research_vehicles}
\end{figure}
Our institute has gained a lot of experience with autonomous driving projects over the last few years.
Starting with the Roborace competition in 2018~\cite{betz2019c}, we developed software capable of autonomous head-to-head racing.
Since then, we have gradually improved our software stack, with up to 30 people working on the stack in parallel.
This enabled us to win the inaugural races of the Indy Autonomous Challenge\,(IAC)~\cite{mitchell2024} and the Abu Dhabi Autonomous Racing League\,(A2RL)~\cite{aspireuae2024}. Fig. \ref{fig:research_vehicles} shows one of our research vehicles: the EAV24 race car we used to compete in A2RL. \\
Additionally, we designed and built a research vehicle called EDGAR, which is tailored to autonomous driving research on public roads~\cite{karle2023}.
This has provided us with valuable insights into evaluating various algorithms across different test vehicles.

While we have shared our general approach in previous publications~\cite{betz2019c,betz2023}, we want to share our insights on combining individual research algorithms into a complete software stack with this work. We derive these insights from practical experience, continuously developing and deploying new research algorithms on vehicles in different racing competitions.
We aim to bridge the gap between research and real-world application by making the following contributions:
\begin{itemize}
    \item Identifying current challenges in developing, maintaining, and deploying autonomous driving software.
    \item Introducing concepts to overcome these challenges.
    \item Validating these concepts using real-world data.
    \item Providing open-source software to implement these concepts at \url{\githuburl}.
\end{itemize}

\section{Related Work}
The Darpa Urban Challenge, held in 2007, is considered a milestone in autonomous driving~\cite {buehler2007}.
From this, various participants presented their approach to developing an autonomous driving software stack~\cite{berger2012,urmson2008,bhat2018}.
However, these publications did not use the Robot Operating System~(ROS)~2, now a de facto standard in autonomous driving research~\cite{reke2020}.
That makes their approaches less applicable in modern software environments.

Today, autonomous racing series offer a comparable challenge to the capabilities of current algorithms. The most prominent are IAC, A2RL, F1TENTH, and Formula Student Driverless (FSD)~\cite{betz2022}. Teams competing in these series often publish their approaches in scientific papers, mainly focusing on algorithmic challenges~\cite{betz2023,raji2024, demeter2024a, okelly2019,alvarez2022,kabzan2019,jung2023,baumann2024}.
On the contrary, \etal{Sauerbeck}~\cite{sauerbeck2023} focus on lessons learned while applying perception systems in high-speed autonomous racing. The paper contains algorithmic challenges and best practices in efficient testing and development.
However, they focus solely on perception systems instead of considering challenges in a full-stack autonomous driving system.
Similarly, \etal{Ögretmen}~\cite{ogretmen2023} focus only on the planning module and its interdependency with the regulations of their racing series.

In \cite{kusmenko2017, hellwig2019}, the authors describe software components using a tailored architecture modeling language. However, their approach was not evaluated
with a system as complex as a full autonomous driving software stack.
\etal{Reke}\cite{reke2020} derived a \ros{} based software architecture for autonomous driving on public roads.
Their main focus was modularity and real-time capability. However, they did their evaluation only on a single test vehicle. Additionally, they did not implement all parts of the proposed architecture before evaluation. Therefore, they could not identify the long-term challenges and disadvantages of their approach.

The open-source project Autoware~\cite{kato2015, kato2018} provides full-stack autonomous driving software.
Autoware has already been deployed in multiple different, even embedded, applications. However, no distinct publications focus on the challenges of using and developing research algorithms with Autoware.
Instead, the annual Autoware Challenges call for solutions to current, predominantly algorithmic problems within the Autoware stack~\cite{mitsudome2024}. Nevertheless, these are just open challenges not addressed by publications so far. Similarly, \etal{Betz}~\cite{betz2022} identify open challenges in autonomous racing without presenting mitigation strategies. \\
In summary, the current state of the art focuses mainly on algorithmic challenges and their influence on software architecture.
No publications to date have identified the current challenges of developing and deploying a complete software stack while also presenting validated solutions to address them within a scientific environment.

\section{Challenges}
\label{sec:challenges}
\begin{figure}[tb]
    \vspace*{0.2cm}
    \centering
    \includegraphics[width=0.8\columnwidth]{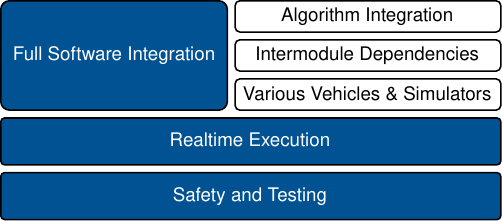}
    \caption{Challenges of developing an autonomous driving software stack.}
    \label{fig:challenges}
\end{figure}

Fig. \ref{fig:challenges} highlights key challenges in developing autonomous driving software, discussed in detail in the following section. These challenges stem from our hands-on experience with research algorithms on real-world test vehicles.

\subsection{Full Software Integration}
\label{subsec:challenge-full_software_integration}
\subsubsection{Algorithm Integration}
\label{subsubsec:challenge-algorithm_integration}
Deploying a research algorithm into an autonomous driving software stack can involve a lot of integration overhead.
This is because research approaches are usually developed and evaluated on public benchmarks~\cite{trauth2024,geisslinger2021,nobis2021}.
However, these benchmarks (e.g., CommonRoad~\cite{althoff2017}, NuScenes~\cite{caesar2019}, KITTI~\cite{geiger2012}) use a custom development environment with a specialized data format.
In contrast, the overall software for autonomous driving is often based on \ros{}~\cite{betz2023, raji2024, kato2018}, which differs heavily from the aforementioned development environments.
In addition, strategies for parameterization and acquiring internal debug variables are required to use the algorithms efficiently. \\
For example, about \SI{50}{\percent} of the open-source vehicle model Open-Car-Dynamics~\cite{sagmeister2024} is not the actual algorithm but the code dedicated to integration.
\subsubsection{Intermodule Dependencies}
Another important aspect that has to be considered is the existence of cross-influences between individual modules. \etal{Baumann}~\cite{baumann2024} stated this as one of their main lessons learned from developing their autonomous racing stack. An example is the interaction between state estimation and control since errors in state estimation inherently influence the control algorithms' performance~\cite{baumann2024}.
This fact is amplified in Global Navigation Satellite System (GNSS) denied environments, where localization is particularly difficult~\cite{goblirsch2024}.
\subsubsection{Adaption to different Simulators and Vehicles}
\esser{
    Additionally, the software must be able to drive many different vehicles, both physical and simulated ones. All previously mentioned autonomous racing series provide a custom simulator so teams can prove their capabilities before real-world deployment. This reduces the probability of crashes and increases safety on track. Since the vehicles differ from competition to competition, the sensor setups, the actuators, and the overall dynamic behavior of the vehicle also vary. Even on the same vehicle, the sensor setup can change from event to event.
    Teams cannot influence the changes made to the hardware since competition organizers determine the hardware for most of the racing series to reduce cost and complexity. \\
    Due to the variety of vehicles and their setups, adapting the autonomous driving software for every setup is time-consuming and error-prone, resulting in different software versions that must be maintained in parallel.
}
\subsection{Real-Time Execution} \label{challenge:realtime_execution}
\betz{
    High-speed autonomous driving requires exceptionally low latency, as signal processing and decision-making delays directly impact vehicle performance~\cite{betz2023analysis}. In addition to the latency of individual algorithms, the total time for data transport between algorithms is important in a full software stack. The resulting cumulative delay from sensor input to actuator output is called end-to-end latency. To ensure seamless operation, the system must execute actions within milliseconds. In addition, predictable timing is crucial. Any variability or jitter in execution can result in inconsistencies that directly undermine competitive performance. \\
    Soft real-time systems, such as those built with \ros{}, provide a practical balance for robotics applications by allowing more timing flexibility for non-critical operations while prioritizing critical tasks. Many algorithms developed in academic research were not designed with real-time adherence in mind. For example, techniques such as dynamic memory allocation, often used in experimental implementations, can introduce unpredictability and violate timing constraints~\cite{reke2020}. Additionally, \ros{} requires careful configuration to meet real-time demands, especially in task scheduling and middleware settings. Furthermore, both high-frequency and high-bandwidth data must be delivered with minimum latency.
    Another layer of complexity comes from using different computing platforms with different performance characteristics. A varying number of cores, clock speeds, and available system memory affect the performance of the executed code and add further constraints. \\
    In addition, data logging is often neglected. Data analytics plays a major role in testing with real vehicles. However, logging the data of the entire autonomous driving software stack consumes a lot of the available hardware resources, possibly interfering with the real-time execution of the software itself.
}
\subsection{Safety and Testing}
\ebner{
    Testing autonomous vehicles without a safety driver poses further challenges that have to be addressed. Testing with prototypes will inevitably lead to faults both in software and hardware. Therefore, a strategy to mitigate these faults safely is essential.
    This task is complicated since most research algorithms are not evaluated concerning degraded or $\text{straight-on}$ missing input signals. Therefore, this usually has to be done during the integration of an algorithm into the software stack, which adds additional overhead.\\
    However, high-speed autonomous racing elevates the need for such a failure mitigation strategy even more. Simple concepts, like emergency braking, cannot be applied safely in all situations, considering the high velocities of such a race car.
    A good safety strategy needs to be able to handle failures of critical sensors at any speed or situation, handle signal degradation smoothly, catch software failures, and finally, choose the required actions to stop or continue with degraded performance. \\
    Choosing the right actions in the right scenarios can prevent losing valuable time by preventing crashes and following repairs. This is amplified by the fact that track time is valuable in full-scale autonomous racing. Testing is usually only possible during official testing periods, where the race track has to be shared among all competing teams.
}

\section{Approaches}

After identifying the challenges in developing a complete autonomous driving software stack, we now present strategies to overcome these challenges.
These strategies are already integrated into the software stack of the TUM Autonomous Motorsport team and have been proven to be successful in practice.
We provide a set of library functions under \url{\githuburl} to reduce the effort to implement the strategies presented in the remainder of this work.

\subsection{Full Software Integration}
\subsubsection{\ros{} Interfaces as a portal into the software}
\begin{figure}[tb]
    \vspace*{0.2cm}
    \centering
    \includegraphics[width=\columnwidth]{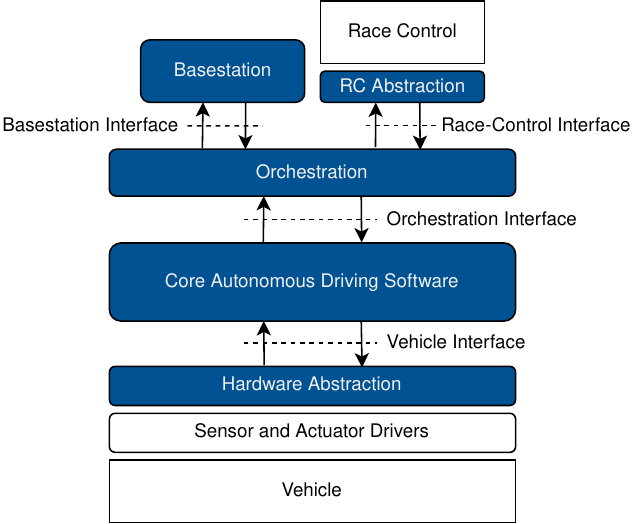}
    \caption{The \ros{} interfaces within our software stack enable abstracted communication.
        The interfaces shown are a fixed list of \ros{} topics with a certain message type and frequency.
        Components depicted in blue are written by the TUM Autonomous Motorsport.}
    \label{fig:ros-interfaces}
\end{figure}

\esser{
    To address the challenge of quickly and reliably adapting to different vehicles, their hardware, and simulators, we have defined fixed \ros{} interfaces for our core software.
}
Fig. \ref{fig:ros-interfaces} shows the main high-level software components and their interfaces. The interfaces are designed so that each module can provide and receive required information in a standardized way. The interfaces are implemented using built-in quasi-standardized \ros{} message definitions. A hardware abstraction isolates the core software and its interface from changes on the vendor (e.g., GNSS receiver) or vehicle side. If the software should be deployed to a new vehicle or simulator, a new hardware abstraction layer has to be implemented and tested. No changes to the core autonomous driving software and the vehicle interface are required.
This approach minimizes maintenance effort and maximizes testing efficiency since the core software remains independent of the vehicle or simulator in which it is deployed.

Additionally, the software stack must receive high-level behavioral commands (e.g., pit, stop, drive slowly). Our participation in the IAC and A2RL requires that the vehicle can be commanded both by the team itself - via the basestation - and race control. Because of this, we created an orchestration layer that uses both these inputs to instruct the core software stack on the desired behavior. The team commands are only allowed for testing purposes and are, therefore, tailored to the specific needs of the team. However, the race control commands fulfill the purpose of racing flags and are defined by race control.
Since race control commands are frequently adapted to the specific race format or racing series, a race control abstraction is implemented that translates those commands to a behavior request (race-control interface). This way, the core software stack is unaffected by changes, and only the abstraction must be adapted and retested.

\subsubsection{Small and Standardized Nodes}

To reduce the integration effort of new algorithms, we took inspiration from software design best practices: the single responsibility principle.
This principle states that a class should only have one reason to change~\cite{martin2003}.
Applying this to our use case means that every \ros{} node should have only a single task. For our stack, this generally meant creating more but smaller nodes instead of a few big ones.

The smaller nodes can be combined more easily to configure the software stack to the specific needs of the vehicle or the track.
Smaller nodes also mean that every module can be developed separately in a development environment and implementation language tailored to the individual algorithm. Further, testing the modules in dedicated pipelines is simplified.
Additionally, we used, wherever possible, either standard \ros{} messages or the message definitions of the open-source autonomous driving stack Autoware~\cite{kato2015, kato2018}.
This way, we can easily integrate different algorithms from Autoware or test our algorithms within the Autoware stack.
This is also beneficial when publishing a developed algorithm since it is designed and implemented from the ground up with Autoware compatibility in mind.

However, having more but smaller nodes increases the number of messages published to exchange information between the nodes. The increased network traffic could lead to problems with message delivery and decrease computational efficiency. Nevertheless, for our software stack consisting of 52 \ros{} nodes, we did not encounter problems or disadvantages from the increased traffic using the measures presented in Sec.~\ref{sec:real-time-execution}\textbf{}.

\subsubsection{Intra-Node Software Design}

With the previously presented strategies, \ros{} nodes are easily interchangeable and can be used in different software stacks.
However, the integration of research algorithms into the \ros{} environment still requires a lot of effort (Sec.~\ref{subsubsec:challenge-algorithm_integration}).

\begin{figure}[tb]
    \vspace*{0.2cm}
    \centering
    \includegraphics[width=0.9\columnwidth]{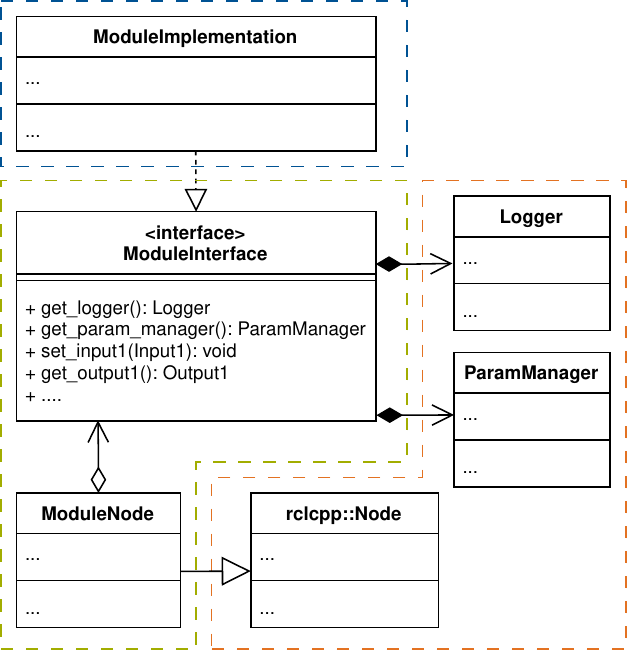}
    \caption{Intra-Node Software Design for reusing a generic \ros{} wrapper to execute different implementations of a certain module. Library components that can be reused among different modules are depicted in orange. Green coloring indicates integration code that can be reused for different implementations of a single module. Blue coloring indicates the implementation of a specific algorithm.}
    \label{fig:in_node_software_design}
\end{figure}

We approach this by strictly separating the algorithm from the \ros{} integration. Fig. \ref{fig:in_node_software_design} shows the internal design of a node resulting from this idea.
We use a strategy design pattern~\cite{gamma1995} to hide all internal details of an algorithm behind a generic interface class. The \ros{} wrapper calls the algorithm through this interface. This way, the algorithm can be easily replaced by different implementations without adapting the \ros{} node. This is especially useful when developing new algorithms since the algorithm can be developed and tested in a different environment (Sec.~\ref{subsubsec:challenge-algorithm_integration}) before being inserted into the \ros{} wrapper.

The base class is derived from the input/output requirements of the wrapping node. However, we keep the interface independent of the \ros{} environment. This is achieved by not using \ros{} messages types within the base class but custom types instead. This way, the base class stays independent, and the derived algorithms can be used in any environment.

With this design strategy, we have the ability to share the integration code among all specific implementations of a certain module. This tremendously reduces the effort to deploy and compare algorithms from the literature.

However, even if all implementations of a certain module share the same inputs and outputs, some things are specific to each algorithm. We have identified the parameters and the logging signals as such algorithm-specific things.
A possible solution is handling these things in a non-standardized way within the algorithm implementation. However, apart from implementing this separately for every algorithm, this comes with additional disadvantages.

\ros{} allows changing parameters during runtime. This powerful feature can drastically improve testing efficiency since testing different parameterizations can be achieved without a potentially time-consuming full software restart. Therefore, we decided to use the parameter API of the \ros{} Node class.
However, to still keep the parameter system of an algorithm independent from \ros{}, we created a zero dependency library that replicates the parameter API of \ros{}. The library provides a parameter manager interface (Fig. \ref{fig:in_node_software_design}) acting as a generic way to access and set parameters from the \ros{} integration. Inside the algorithm, we just declare which parameters are required, and the integration code is responsible for setting the correct values and keeping them up to date. Additionally, having a common API to set parameters among all modules allowed us to develop several utility functions, e.g., validating that all parameter overrides from a file have been set correctly.

Similarly, logging signals inside the algorithm implementation is possible. However, the high number of nodes in the software stack would result in many scattered log files, which complicates debugging full-stack problems. Additionally, ensuring safety and real-time capability when writing to the file system imposes additional programming overhead. \\
For these reasons, we decided to use rosbag as a logging format. It allows comprising the regular \ros{} messages and internal log signals in a single file.
It simplifies analyzing algorithm performance and possible implementation errors since all relevant information is within a single file. Additionally, rosbags can be easily replayed for simulation, a powerful feature during development.
However, logging signals are specific to each algorithm, which does not allow for a generic \ros{} wrapper as introduced above. Furthermore, internal logging signals usually change during the development of an algorithm. Thus, regularly modifying a debug message definition specific to the algorithm induces overhead and breaks compatibility with previously recorded rosbags.
Therefore, we designed a zero-dependency library that collects debug values inside an algorithm and publishes them in \ros{}. Instead of using conventional \ros{} message definitions, we serialize all information into arrays of numeric values. By additionally publishing the scheme containing the signal names, these can be deserialized to restore the original values, effectively allowing for a dynamic \ros{} message definition.
We provide an open-source implementation available in both C++ and Python with this publication under \url{https://github.com/TUMFTM/tsl}. To improve usability, we integrated automatic parsing of the logged signals in the open-source time-series analysis tool PlotJuggler~\cite{davidefacontiPJ}.

\subsection{Real-Time Execution}
\label{sec:real-time-execution}
\betz{
    After improving the efficiency of integrating the algorithm into the software stack, we present three aspects to tackle the challenges of executing the created node with certain real-time requirements.
}
\subsubsection{Latency Measurement}
\betz{
    Maintaining low latencies in the autonomous driving software stack is crucial for real-time performance. This requires continuous monitoring and accurate measurement of execution and communication times. The \texttt{ros2\_tracing} framework~\cite{bedard2022ros2_tracing} provides tools specifically designed for measuring \ros{} applications. The framework enables accurate and low overhead latency measurements by inserting trace points directly into the \ros{} client library \texttt{rclcpp}.
    A dedicated framework~\cite{betz2023latency} has been implemented to get even deeper insights. This framework is capable of evaluating not only the overall latency but also the jitter of periodic callbacks. While this approach provides detailed insights, the extended conversion times for tracing data can hinder the timely debugging of timing violations. To overcome this limitation, message stamping is enforced for each publisher. It allows for efficient post-analysis of latency issues using the recorded rosbags, significantly improving the analysis workflow.
    By continuously monitoring both execution times and end-to-end latencies, runtime bottlenecks in the software stack can be identified. This insight facilitates targeted optimization efforts.
    In addition, every node monitors all of its input topics strictly for timeouts during operation. For details see Sec.~\ref{subsec:safety_concept}.
}
\subsubsection{\ros{} Scheduling \& Optimization}
\betz{
    Effective scheduling of tasks is essential to ensure the real-time performance of \ros{} systems, especially in demanding applications such as high-speed autonomous driving.
    CPU isolation and pinning, effectively dedicating specific CPU cores to high-priority tasks, are established methods. By isolating critical processes from non-critical workloads, CPU contention is minimized, and timing predictability is improved. In our case, the most critical processes are trajectory control, state estimation, and the drivers required to actuate the car. The isolation ensures these tasks can run uninterrupted and maintain their real-time guarantees. Also, interrupt handling for the individual cores must be configured in the kernel.
    Best practices for real-time execution state that hyper-threading should be avoided to improve timing consistently. This is also true for \ros{} based software~\cite{betz2023fast}.
}

\betz{
    Using \ros{} for big software projects is challenging due to its asynchrony and configuration complexity. While writing individual \ros{} nodes is a simple task, ensuring the performance of an entire \ros{} system designed to run \SI{75}{\meter\per\second} requires careful design.
    We, therefore, analyzed the whole system regarding the best \ros{} configuration parameters~\cite{betz2023fast, teper2023timing}. Accordingly, using message queue sizes of length one is a simple yet effective measure. This reduces the latency as each module only processes the latest received data (best effort). Additional influencing factors in \ros{} are the timer periods, node executor assignment, DDS mode, executor and node priorities, node composition, and intra-process communication.
    We designed an optimization using an analytical latency model to optimize these influences for our software stack~\cite{teper2024end}.
    Using this framework and the aforementioned measures, we successfully tackle the problem of real-time execution of our \ros{} software stack.
}

\begin{figure}[tb]
    \vspace*{0.2cm}
    \centering
    \includegraphics[width=0.9\columnwidth]{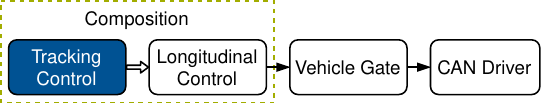}
    \caption{An exemplary high-frequency communication chain within the software of TUM Autonomous Motorsport. Nodes depicted in blue publish based on a timer callback. Nodes depicted in white are directly triggered by the message subscription. The two control nodes are composed into one executor and communicate utilizing intra-process communication.}
    \label{fig:control-chain-latency}
\end{figure}

\begin{figure}[tb]
    \centering
    \includegraphics{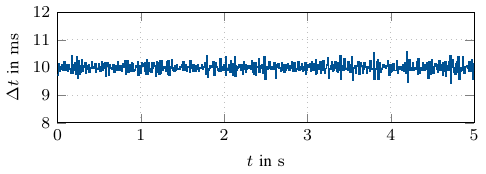}
    \vspace*{-0.2cm}
    \caption{Time difference between two consecutive CAN frames belonging to the communication chain depicted in Fig. \ref{fig:control-chain-latency}. The data originate from the EAV24 race car during an A2RL testing session.}
    \label{fig:heartbeat-consistency}
\end{figure}

Fig. \ref{fig:control-chain-latency} shows a communication chain in our software well suited for demonstrating execution and message delivery timing.
To achieve low latency, only a single node publishes with a timer callback, while all other nodes directly republish on subscription after the required internal calculations.
In such a scenario, the potential jitter accumulates over all nodes in the chain.
\pagebreak
Fig. \ref{fig:heartbeat-consistency} shows that our approach, even in this case, keeps the maximum deviation from the reference interval lower than \SI{0.57}{\milli\second} while showing a standard deviation of \SI{0.17}{\milli\second}.

\subsection{Safety Concept}
\label{subsec:safety_concept}

\subsubsection{Module-Level Measures}\label{sec:strategy_watchdog}

Our modular approach also transcends into our safety concept. Each module is responsible for determining its state and reporting back cyclically to the rest of the software. Therefore, each module has to watch for missing, corrupted, or invalid information as well as message timeouts and report its state accordingly. On top of reporting its internal state, every module is responsible for taking the correct measures to appropriately react to internal and external faults. An example is the control module, which independently switches to the emergency trajectory when it no longer receives valid trajectories.

We use \ros{} diagnostic messages to report the internal states of the individual modules.
The diagnostics framework can deliver various information that helps to determine the cause of a problem. However, we decided to only use the diagnostic level for runtime decisions in the orchestration. The diagnostic level comprises the four levels \stateok{}, \statewarn{}, \stateerror{} and \statestale{}. However, each module can provide additional information in the diagnostic message, e.g., the reason for the error, which is used for debugging and online visualization in the basestation.

To reduce the implementation effort, we designed and implemented library functions to uniformly monitor subscriptions for timeouts and publish the internal module state.

\subsubsection{Orchestration-Level Measures}\label{sec:strategy_state_machine}

The orchestration is responsible for aggregating and monitoring the states of the individual modules and consists of a watchdog and a state machine. \\
The watchdog assembles the individual asynchronous diagnostics for all the modules into a single software state report.
It also overwrites the last received module state with \statestale{} in case a module crashes and stops reporting its state.
Based on this cyclically published software state, the state machine computes one of the following actions:
\ebner{
    \begin{itemize}
        \item Nominal operation: no further action
        \item Safe stop: target speed is set to zero. The vehicle is slowing down to a full stop
        \item Emergency stop: switch to the emergency trajectory using maximal deceleration
        \item Hard emergency: apply maximum brake pressure and steer straight.
    \end{itemize}
}
\ebner{
    Safe stops are triggered on non-critical faults, like communication dropouts between vehicle and base station and failure of non-critical software modules. The vehicle decelerates safely, with all remaining software components being active.
}

\ebner{
    Emergency stops are triggered by no longer receiving valid trajectories or failure to track the performance trajectory. These emergency trajectories are calculated simultaneously with the target trajectory and decelerate to a standstill. The controller always stores the last valid trajectory that was received. By doing so, we ensure a valid stopping trajectory in case of a non-recoverable failure of the planning module at any instant. While following the emergency trajectory, no newly calculated trajectories are accepted. Thus, it is not possible to further react to the dynamic environment~\cite{ogretmen2023}.
}

The hard emergency is a last-ditch effort within the software stack to bring the vehicle to a stop in case of critical failure of software modules and sensors.
During a hard emergency, we apply a predefined brake pressure and straighten the steering to decelerate as hard as possible while aiming to go straight.
In the current software stack, only a crash of highly critical modules (control, state estimation), a big lateral offset to the reference trajectory, or failure to localize will trigger a hard emergency.
Hard emergencies are facilitated by the vehicle gate, which runs in a process different from the control module to harden the system against potential crashes of the control module.

\begin{figure}[tb]
    \vspace*{0.2cm}
    \centering
    \includegraphics[width=0.8\columnwidth]{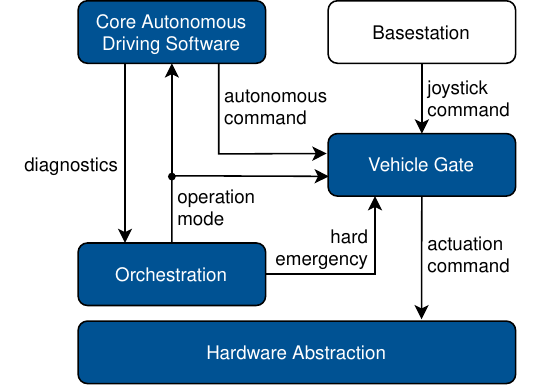}
    \caption{Safety Architecture. The components depicted in blue are running on the test vehicle itself.}
    \label{fig:safety_architecture}
\end{figure}

Two conditions can activate the hard emergency mode in the vehicle gate. This is either a timeout of the actuation commands from the autonomy or a direct instruction by the state machine (Fig. \ref{fig:safety_architecture}).

In addition, if a module cannot provide meaningful outputs anymore, it reports a \statestale{} status to the state machine while simultaneously stopping to publish.
This way, there is no distinction between an outright crash and a module not being functional anymore. By actively reporting a \statestale{}, we improve reaction times compared to relying on the watchdog to notice a timeout on the status message.

\begin{figure}[tb]
    \vspace*{0.2cm}
    \centering
    \includegraphics{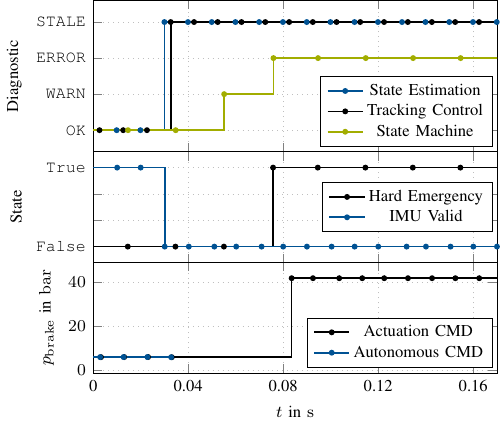}
    \caption{Hard Emergency triggered by losing all available IMUs during testing on the Yas Marina Formula 1 Track in Abu Dhabi.}
    \label{fig:plot_he}
\end{figure}

Fig. \ref{fig:plot_he} shows such a hard emergency situation during real-world testing. After a crash of the GNSS and Inertial Measurement Unit (IMU) driver, localization was no longer possible, which was indicated by the state estimation reporting a \statestale{} status. Because every module checks the validity of its inputs, the controller also switched to \statestale{} immediately afterward to indicate that vehicle control is impossible without a valid localization. Since modules stop publishing while being in \statestale{} state, there was no longer an autonomous command.
The state machine cycles with a timer period of \SI{20}{\milli\second}. Therefore, and because the watchdog must assemble the diagnostics first, the state machine's reaction to the localization failure was delayed. Due to an implementation error, the state machine took two cycles to reach the appropriate \stateerror{} state, which triggers the hard emergency request to the vehicle gate.
Immediately following the hard emergency request, the vehicle gate requested the desired brake pressure of more than \SI{40}{\bar}. Fig. \ref{fig:plot_he} proves that, without the implementation error, this concept would have been able to apply the brakes within approximately \SI{20}{\milli\second} after the first module failure while maximizing modularity and adaptability. %

\subsubsection{Manual Intervention}\label{sec:strategy_manual_intervention}

As a last resort, the longitudinal control command can be overridden by the joystick operator during the autonomous operation of the vehicle.
This is also handled by the vehicle gate (Fig. \ref{fig:safety_architecture}) and allows cutting the throttle and applying the brakes.
This function can be used to quickly stop the vehicle, e.g., in case of a failure to detect other cars or when the car visibly misbehaves.
Additionally, the vehicle gate offers a manual driving mode. In this mode, we directly control the throttle, brakes, and steering via the joystick operator.
\ebner{Even though manual intervention is not a replacement for a safety driver and is rarely used during testing, it is useful to quickly resolve issues that would otherwise require a towing vehicle. This improves testing efficiency by saving important time on track.}

\section{Conclusion}
\label{sec:conclusion}

In this work, we derived challenges and strategies involved in developing a full autonomous driving software stack. Drawing from our practical experience, we identified key difficulties in integrating research algorithms into a full-stack system and deploying them on real-world test vehicles. We introduced several strategies to overcome exactly these challenges, including modular software design, the use of abstracted \ros{} interfaces for communicating out of the stack, and robust safety measures. Our approaches have been successfully implemented in our software, enabling us to achieve several milestones in autonomous racing competitions, such as winning the inaugural IAC and A2RL races. Furthermore, we provide open-source implementations of our solutions to facilitate future research and development in the field of autonomous driving. By sharing our insights and tools, we aim to simplify the deployment of newly developed algorithms into full software stacks. This is tremendously important for evaluating the real capabilities of certain algorithms as we have outlined in Sec.~\ref{sec:challenges}.
With the contributions of this work, we bridge the gap between research and real-world application, ultimately advancing the state of autonomous driving technology.

\section*{Acknowledgment}
Author contributions: Simon Sagmeister, as the first author,
designed the structure of the article and contributed essentially to identfying the current challenges, designing and implementing the solution approaches, and the overall contents.
Simon Hoffmann, Tobias Betz, Dominic Ebner, and Daniel Esser contributed both to the implementation of the presented concepts as well as the contents of the research paper.
Markus Lienkamp made an essential contribution to the concept of the research project. He revised the paper critically for important intellectual content. Markus Lienkamp gives final approval for the version to be published and agrees to all aspects of the work. As a guarantor, he accepts responsibility for the overall integrity of the paper.

\bibliographystyle{IEEEtran}
\bibliography{IEEEabrv, bibliography/stylecustomization.bib, bibliography/literature.bib, bibliography/literature_betz.bib, bibliography/literature_ebner.bib, bibliography/literature_esser.bib}

\end{document}